# Knowledge Engineering for Large Belief Networks

**Malcolm Pradhan**[1]  **Gregory Provan**[2]
[1]Section on Medical Informatics, MSOB X-215,
Stanford University, CA 94305

**Blackford Middleton**[1]  **Max Henrion**[2]
[2]Institute for Decision Systems Research
4894 El Camino Real, Suite 110, Los Altos, CA 94022

## Abstract

We present several techniques for knowledge engineering of large belief networks (BNs) based on the our experiences with a network derived from a large medical knowledge base. The noisy-MAX, a generalization of the noisy-OR gate, is used to model causal independence in a BN with multivalued variables. We describe the use of leak probabilities to enforce the closed-world assumption in our model. We present Netview, a visualization tool based on causal independence and the use of leak probabilities. The Netview software allows knowledge engineers to dynamically view subnetworks for knowledge engineering, and it provides version control for editing a BN. Netview generates sub-networks in which leak probabilities are dynamically updated to reflect the missing portions of the network.

## 1 INTRODUCTION

Given the relative maturity of algorithm development in the Bayesian reasoning community, attention is now starting to focus on applying these algorithms to real-world applications. The Quick Medical Reference–Decision Theoretic (QMR–DT) project seeks to develop practical decision analytic methods for large knowledge-based systems. The first stage of the project converted the Internist-1 knowledge base [Miller, Pople et al. 1982] (QMR's predecessor) into a binary, two-layered belief network (BN) [Middleton, Shwe et al. 1991; Shwe, Middleton et al. 1991]. In the second stage of the QMR-DT project we are creating a multilayer belief network with multivalued variables, and developing efficient inference algorithms for the network. This paper will concentrate on the knowledge engineering issues we faced when building a large multilayered BN.

To create a large multilevel, multivalued BN we took advantage of a rich knowledge base, the Computer-based Patient Case Simulation system (CPCS–PM), developed over two years by R. Parker and R Miller [Parker and Miller 1987] in the mid-1980s as an experimental extension of the Internist-1 knowledge base.

This paper makes contributions both in knowledge engineering and in algorithm development and implementation for large BNs. We describe the CPCS BN that we developed, the dynamic network tool that we designed and implemented to aid knowledge engineering for CPCS, and the Bayesian network algorithms that we employed for this large, complex network.

The CPCS is one of the largest BNs in use at the current time, and its sheer size makes most tasks, such as knowledge engineering or evaluation, challenging. The development of CPCS necessitated the implementation of algorithms that could make inference in BNs of this size more efficient. An example of this is a generalization of the *noisy-OR* gate [Cooper 1986; Peng and Reggia 1987; Pearl 1988] that is commonly used in binary valued networks to model causal independence. The CPCS BN contains nodes that are multivalued, for example, disease nodes may have four values (absent, mild, moderate, severe), consequently we use a generalization of the noisy-OR gate called the *noisy-MAX*. The specification of a complete conditional probability matrix for a node $m$ with $s_m$ values and $n$ predecessors requires the assessment of $(s_m - 1)\prod_{i=1}^{n} s_i$ probabilities, where $s_i$ is the number of values of predecessor $i$ (for a binary network this reduces to $2^n$). In contrast, the causal independence assumption in the form of a noisy-gate reduces this assessment task to $\sum_{i}^{n}(s_m - 1)s_i$ probabilities. thereby simplifying knowledge acquisition and greatly reducing storage requirements.

To aid in the editing and refinement of the CPCS BN, we have developed a network visualization tool we named *Netview*. The Netview tool provides dynamic views of the BN, and can generate subnetworks by taking advantage of the noisy-MAX and leak assumptions. For inference, the network, or subnetworks generated by Netview, are sent to the IDEAL package [Srinivas and Breese 1989] for inference. Netview is a flexible tool which can be used in any BN that uses noisy-gates, and is described in section 5.1.

Like the Internist-1-derived BN, the CPCS BN uses *leak probabilities* [Henrion 1988] to represent the chance of an event occurring when all of its modeled causes are absent. We discuss our use of leak probabilities, and the modifications to the leak probabilities required by the dynamic network tool, in section 5.2.

## 2 KNOWLEDGE BASE TO BELIEF NETWORK

The CPCS–PM system is a knowledge base and simulation program designed to create patient scenarios in the medical sub-domain of hepatobiliary disease, and then evaluate medical students as they managed the simulated patient's problem. Unlike that of its predecessor Internist-1, the CPCS–PM knowledge base models the pathophysiology of diseases—the intermediate states causally linked between diseases and manifestations. The original CPCS–PM system was developed in FranzLisp. Diseases and intermediate pathophysiological states (IPSs) were represented as Lisp frames [Minsky 1975].

To construct the BN we converted the CPCS–PM knowledge base to CommonLisp and then parsed it to create nodes. We represented diseases and IPSs as four levels of severity in the CPCS BN—absent, mild, moderate, and severe. Predisposing factors of a disease or IPS node were represented as that node's predecessors, and findings and symptoms of a disease or IPS node as the successors for that node. In addition to the findings, CPCS contained causal links between disease and IPS frames, we converted these links into arcs in the BN. Frequency weights [Shwe, Middleton et al. 1991] from the CPCS–PM ranged from 0 to

5 and were mapped to probability values, as described in the next section. Frequency weights from the CPCS–PM were mapped to probability values based on previous work in probabilistic interpretations of Internist-1 frequencies.

We generated the CPCS BN automatically, we did manual consistency checking using domain knowledge to edit the network. Because the CPCS–PM knowledge base was not designed with probabilistic interpretations in mind, we had to make numerous minor corrections to remove artifactual nodes, to make node values consistent and to confirm that only mutually exclusive values were contained within a node.

The resultant network has 448 nodes and 908 arcs (Figure 1). A total of seventy four of the nodes in the network are predisposing factors and required prior probabilities, the remaining nodes required leak probabilities assessed for each of their values. We thus had to assess over 560 probabilities to specify the network fully.

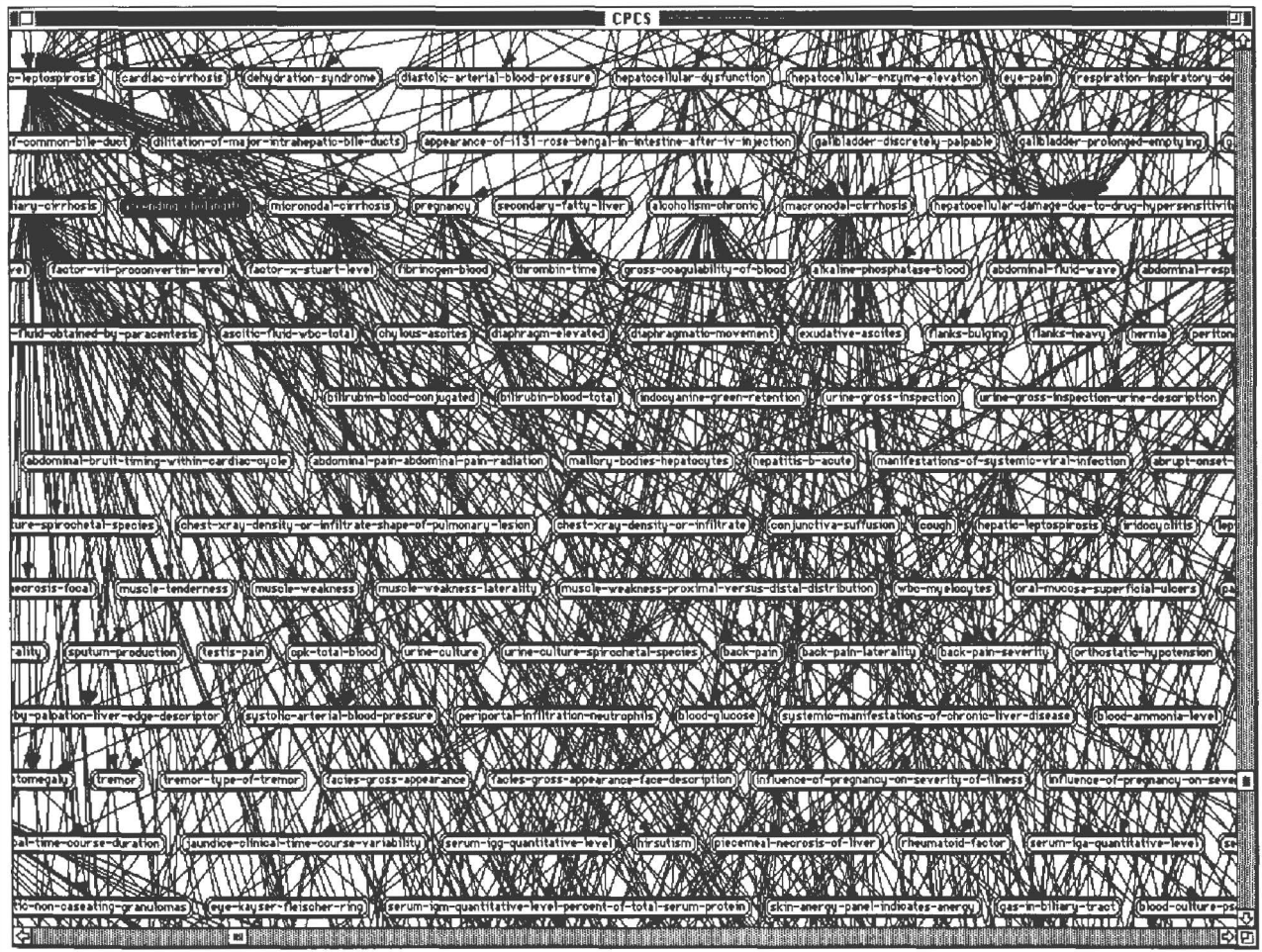

Figure 1. A small portion of the CPCS BN displayed in the Netview visualization program. The node *ascending-cholangitis* in the third row shown in inverse has been selected by the user.

While the CPCS-PM knowledge base is derived from the Internist-1 knowledge base it has been significantly augmented by inclusion of the IPS states, and multivalued representations of both diseases and manifestations of disesase. The original QMR-BN transformation of the Internist-1 knowledge base used only binary valued disease and manifestation nodes. While conceptually simple, this approach does not adequately reflect the potential variation in presentation of disease manifestations, or the severity of diseases.

## 3    GENERALIZATION OF THE NOISY-OR

### 3.1    NOISY-OR

The noisy-OR is a simplified BN representation that requires far fewer parameters than the full conditional-probability matrix. The noisy-OR is defined over a set of binary-valued variables, and is typically described for a two level network partitioned into two sets of variables, which are interpreted as either cause and effect variables respectively, or disease and manifestation variables respectively [Peng and Reggia 1987]. Consider an effect variable $X$ that has $n$ cause variables or predecessors $D_1,...,D_n$. The noisy-OR can be used when (1) each $D$ has an activation probability $p_i$ being sufficient to produce the effect in the absence of all other causes, and (2) the probability of each cause being sufficient is independent of the presence of other causes [Henrion 1988].

In this paper, we define variables using upper-case letters, and values that variables can take on using lower-case letters. The domain of possible values for variable $X$ is $\Omega_X$. If variable $X$ is present it is denoted using the letter $x$; if it is absent, it is denoted using $\bar{x}$.

The activation of a variable $X$ by a predecessor variable $D_i$ is independent of the value of $D_i$. Under the noisy-or assumption, $X$ is activated when $D_i$ is active, with a conditional probability given by $p_i = P(x \mid d_i \underset{k \neq i}{\wedge} \bar{d}_k)$. In other words, this activation probability denotes the probability when $D_i$ is active and all other predecessors are inactive.

For a two-level noisy-OR network, we define a set $V$ of cause or disease variables, and a set $W$ of effect or manifestation variables. If there is a set $V_I$ of $V$ of predecessors of $X \in W$ that are present, then since the $D_i$ in $V_I$ are independent, $X$ will be false when *all* $D_i$ are false:

$$P(X = \bar{x} \mid V_I) = \prod_{i: D_i \in V_I} P(D_i = \bar{d}_i)$$

$$P(X = \bar{x} \mid V_I) = \prod_{i: D_i \in V_I} (1 - p_i).$$

From this, it is simple to derive

$$P(X = x \mid V_I) = 1 - \prod_{i: D_i \in V_I} (1 - p_i).$$

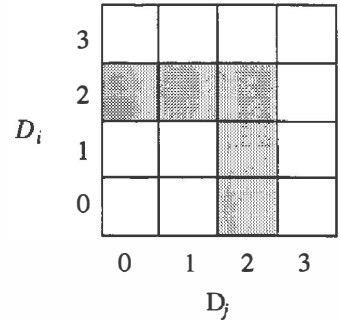

Figure 2. A node $x$ with predecessors $D_i, D_j \in V_I$ having ordered states $\{0,1,2,3\}$. The shaded area represents the probabilities required to calculate the $P(X= 2 | V_I)$.

If there are multiple "manifestation" variables $X_j$, for $j=1,...,l$, then we obtain

$$P(X_j = x_j \mid V_I) = 1 - \prod_{i: D_i \in V_I} (1 - p_{ij})$$

where $p_{ij}$ is the link probability on the arc from $D_i$ to $X_j$.

The simple noisy-OR is insufficient for the CPCS BN application, since we need to accommodate $n$-ary variables. For example, CPCS BN disease and IPSs can take on values such as absent, mild, moderate, and severe.

### 3.2    NOISY-MAX

Consider a generalization of the noisy-OR situation in which each variable is allowed to have a finite discrete state space (rather than just a binary state space). This generalization was first proposed by [Henrion 1988], but he did not describe the algorithmic details. In developing this generalization, we assume that we have a set $V$ of predecessor variables $D_1,...,D_n$. Consider first the case where we have a variable $X$ with a subset $V_I$ of $V$ that are present, with the predecessors indexed by $i,j,...,q$.

The variable domains in CPCS BN are all partially ordered, for example, {absent, mild, moderate, severe}, and it turns out that such a partial ordering is necessary for all variable domains. For the remainder of our work we assume that all variables have ordered domains.

We now want to compute

$$P(X = x \mid V_I) = P(D_i = d_i, D_j = d_j,..., D_q = d_q).$$

The value $x$ is given by $x = \max(i, j,...,q)$ [Henrion 1988]. In other words, $X$ takes on as its value the maximum of the domain values of its predecessors, given that the predecessors are all independent.

This computation of $P(X = x \mid V_f)$ can be viewed as setting up a hypercube of dimensions $i \times j \times \cdots \times q$ and summing the cells, each of which contains a probability value $P_{ijk\cdots q} P_{ijk\cdots q}$. As an example of the derivation of the general formula, we consider the case of two predecessors $D_i$ and $D_j$. If these variables take on values $i, j$ respectively, then the probability $P(X = x \mid V_f)$, where $x = \max(i, j)$. For example, Figure 2 graphically depicts the conditional probability matrix for $D_i$ and $D_j$, both of which have ordered states $\{0,1,2,3\}$. If $x=2$, then $P(X = 2 \mid D_i, D_j)$ consists of the shaded squares of the grid.

In this multivalued noisy-MAX situation, the probabilities that are being calculated in these hypercubes are cumulative probabilities, that is, $P(X \le x \mid D \le d)$. For notational convenience, we define the cumulative probability for a variable $X$ that has a single predecessor $D$ with maximum domain value $d$ as:

$$\Psi(x \mid d) = P(X \le x \mid D \le d).$$

Under the generalized noisy-OR assumption, for a given variable $X$ with a set of $q$ predecessors $D_1, \ldots, D_q$ for which each $D_i$ has maximum value $d_i$, we know that

$$\Psi(x \mid d_i, i = 1, 2, \ldots, q) = \prod_{i:D_i \in V_i} P(X \le x \mid D_i \le d_i)$$
$$= \prod_{i:D_i \in V_i} \Psi(x \mid d_i).$$

Note that using this transformation, we can define the probability assigned to $X$ taking on a particular value $x_k$:

$$P(X = x_k \mid D_i) = \begin{cases} \Psi(x_k) - \Psi(x_{k-1}) & \text{if } x_k \ne 0 \\ \Psi(0) & \text{if } x_k = 0 \end{cases}$$

The unconditional probability assigned to a variable can be derived in an analogous fashion. First, we need to derive the unconditional probability of variable $D_i$, assuming no predecessors. As described in [Provan 1994], this is given by

$$P(X \le x) = P(L \le l) \prod_{i:D_i \in V_i} [p_i P(D_i \le d_i) + (1 - p_i)] P(x \mid V_i).$$

The unconditional probability is given by

$$P(X \le x) = \prod_{i:D_i \in V} [p_i P(D_i \le d_i) + (1 - p_i)].$$

The unconditional probability assigned to X taking on a particular value x is:

$$P(X = x_k) = \begin{cases} P(X \le x_k) - P(X \le x_{k-1}) & \text{if } x_k \ne 0 \\ P(X = 0) & \text{if } x_k = 0 \end{cases}$$

Using this approach, the value $P(x \mid V_f)$ can be computed in time proportional to the number of predecessors in $V_f$. This generalized noisy-MAX has been implemented in IDEAL.

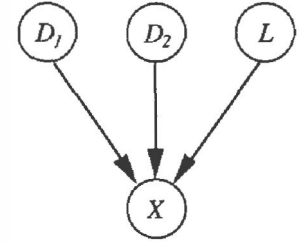

Figure 3. Explicit representation of the leak probability as a cause of $X$.

## 4  LEAKS

As in any other knowledge representation scheme, the BN representation suffers from incompleteness, in that it typically cannot model every possible case. A leak variable can be used to enforce the closed-world assumption [Reiter 1978]. The leak variable represents the set of causes that are not modeled explicitly. A *leak probability* is assigned as the probability that the effect will occur in the absence of any of the causes $D_1, \ldots D_n$ that are explicitly modeled. If the leak variable is explicitly modeled, then it can be treated like any other cause and can be depicted as such (Figure 3). In this representation, the leak node is always assumed to be "on", that is $P(L=true) = 1.0$.

If the leak $L$ with value $l$ is factored into the calculation of the unconditional probability for variable $X$, then we obtain

$$P(X \le x) = P(L \le l) \prod_{i:D_i \in V} [p_i P(D_i \le d_i) + (1 - p_i)],$$

Explicitly representing leak nodes in the CPCS BN would almost double the size of the network, so leaks are implicitly represented in the probability tables of a node's conditioning parents. The Netview knowledge engineering tool, described in section 5, facilitates the maintenance and editing of leak probabilities. Netview stores leak values separately, as if they were explicit nodes, and then inserts the leak values into the appropriate probability tables when exporting a network for inference in IDEAL.

## 5  TOOLS FOR KNOWLEDGE ENGINEERING

### 5.1  NETVIEW: A TOOL FOR NETWORK VISUALIZATION AND EDITING

Verification and refinement of the structure of the CPCS BN is an important part of the QMR-DT project for two reasons. First, because the CPCS BN was generated from a pre-existing knowledge base. Second, the effect of model structure on network performance and accuracy is an important aspect of the QMR-DT project.

During the knowledge engineering process, it became obvious that available tools were not suitable for visualizing and editing a network the size of CPCS BN (Figure 1). In particular, most tools only permit a static view of the network, a limitation that made editing the CPCS network very hard.

Netview was created to help knowledge engineering efforts by allowing knowledge engineers to focus on portions of the network. The program is implemented in Macintosh CommonLISP 2.01. The main features of Netview are

- dynamic network layout
- semantic labeling of nodes
- version control
- subnetwork generation and dynamic leak modification
- leak value editing

Because of the causal independence assumptions implied by the use of the noisy-M A X and noisy-O R gates, knowledge engineers are can select smaller parts of the CPCS BN for viewing. Netview allows the user to view all ancestors, all predecessors, or all ancestors and predecessors of selected nodes. For example, in Figure 1 while the node *ascending-cholangitis* is selected (inverse color), we can use Netview's ability to show all successors and predecessors of the selected node or nodes, resulting in the subnetwork view shown in Figure 4. Other options include viewing nodes' Markov blanket, and immediate successors or predecessors.

Netview uses a dynamic layout algorithm to display the selected nodes. The knowledge engineer is able to move rapidly between views by selecting nodes and choosing viewing options, or by retrieving previously saved views. Quickly viewing a node's predecessors allows rapid assessment of leak probabilities.

In addition to subnetwork selection, Netview allows semantic labeling of nodes, and filtering based on semantic labels. For example, nodes in CPCS BN are labeled "lab finding," "symptom," "sign," "disease," "IPS," "liver disease," and so on. A node may have any number of semantic labels. Semantic labeling is a useful technique for filtering nodes to focus attention during knowledge engineering. It is possible, say, to focus only on "gastric" findings and diseases when dealing with the appropriate domain expert. In the future we will also use the semantic labels in the dynamic layout algorithm to improve the appearance of subnetwork views.

It is useful to keep track of modifications while editing the BN. To facilitate this, Netview includes basic version control to store deleted and added arcs and nodes and changes to probability tables. Arc and node additions and removals between versions are displayed through the use of different colors.

## 5.2    SUBNETWORK GENERATION AND DYNAMIC LEAK MODIFICATION

### 5.2.1    Subnetwork generation

The Netview program is used only for network visualization and editing; Netview saves files in IDEAL format for inference. Because of the size of the CPCS BN it is not always desirable to send the entire network to IDEAL for inference. If we are only interested in verifying a small set of diseases we can generate a subnetwork including only those diseases of interest and their associated findings, IPSs and predisposing factors. When we run test cases against a subnetwork we don't require the system to compute the posterior probabilities of diseases that we are not interested in.

### 5.2.2 Dynamic leak modification

Subnetworks we select from the full CPCS BN using Netview can be exported to IDEAL for inference. It is possible to select subsets of the larger CPCS BN for inference due to the presence of leaks.

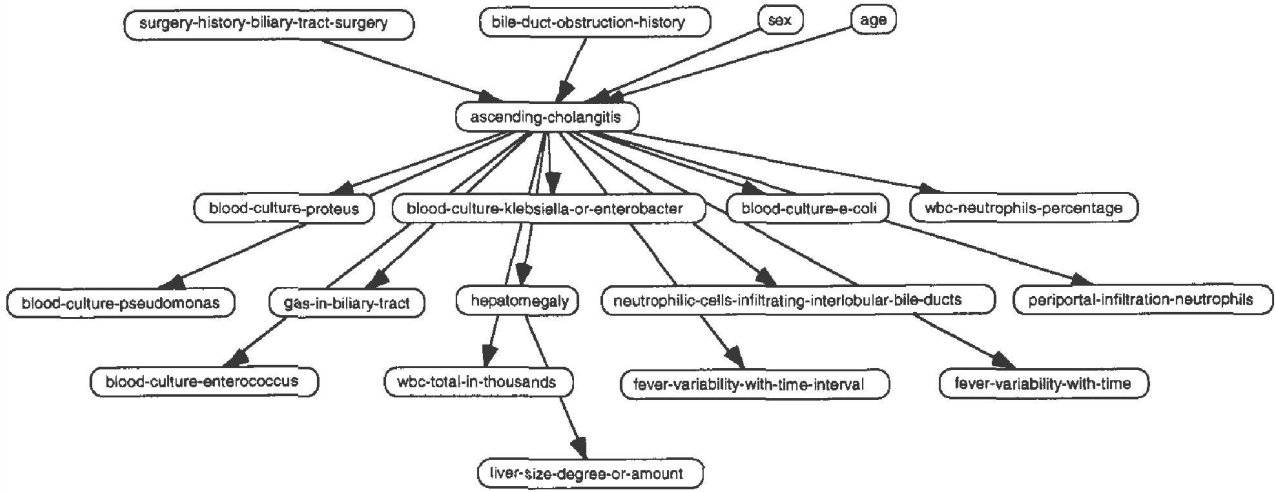

Figure 4. A subnetwork of the CPCS BN displayed in Netview. This view comprises all predecessors and successors of the node *ascending-cholangitis*.

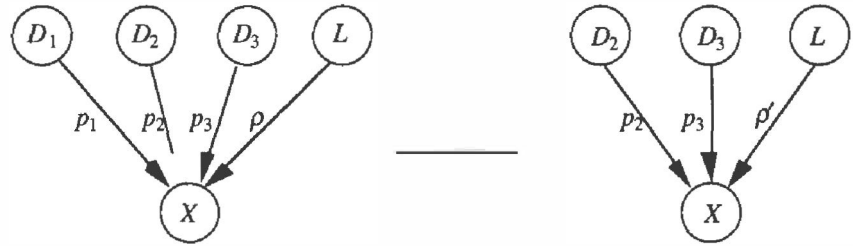

Figure 6. Subnetwork creation. Node $D_3$ is removed from the network, the value of the leak node $l, \rho$, is updated to $\rho'$ based on the probability of $D_3$.

When a subnetwork is saved Netview updates the leak probabilities to take into account the missing diseases. In the CPCS BN the node *hepatomegaly* has parents shown in figure 5, The leak probability for *hepatomegaly* is therefore calculated based on this set of predecessors. In figure 4 a subnetwork was selected based on the ancestors and predecessors of the disease *ascending-cholangitis*. Consequently, the only parent of *hepatomegaly* in the subnetwork is *ascending-cholangitis*, its other parents are not included. The transformation of leak probabilities required during subnetwork creation is shown in Figure 6. The leak probability must be updated from $\rho$ to $\rho'$. This updating is done in order to preserve the total probability mass. If the value $l$ of $L$ is updated to a value $l'$ for new leak $L'$, we can compute the updated leak node probability as

$$\rho' = P(L \leq l \wedge D_3 \leq d_3)$$
$$= P(L \leq l)P(D_3 \leq d_3)$$
$$= [p_3 P(D_3 \leq d_3) + (1 - p_3)]$$

If we want to combine a set of $Q$ nodes into a leak node, where each node $d_i$ in $Q$ has link probability, then the new leak node probability is given by:

$$\rho' = P(L \leq l \wedge D_1 \leq d_1 \wedge \cdots \wedge D_q \leq d_q)$$
$$= P(L \leq l) \prod_{i:D_i \in Q} P(D_i \leq d_i)$$
$$= P(L \leq l) \prod_{i:D_i \in Q} [p_i P(D_i \leq d_i) + (1 - p_i)].$$

We prove in [Provan 1994] that if the network is hierarchical and there are no arcs between nodes at the same level of the hierarchy, then the leak updating is sound, that is, the probability assigned to $X$ given the new set of predecessors is the same as the probability assigned to $X$ with the original predecessors. This proof holds if the subnetwork consists of a Markov blanket of a node, all predecessors and successors of a node, or all successors of a node. The assumption for the proofs holds for the CPCS

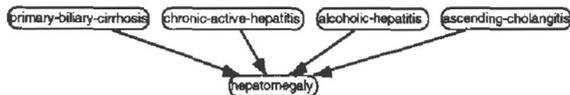

Figure 5. Parents of the node *hepatomegaly*.

BN, and we are exploring how much the network performance changes when the assumption is relaxed.

### 5.2.3 Information metrics

When subnetworks are created some information is lost as parts of the network are excluded. A future area of research is to use Netview to calculate the information loss of a subnetwork based on information metrics [Provan 1993], and to compare differences in posterior probability between the complete network and the subnetwork which has been selected.

## 6  RELATED WORK

The generalization of the noisy-OR was first proposed in [Henrion 1988], and the derivation and implementation described here follow that original proposal. Two related generalizations are described in [Srinivas 1993] and [Diez 1993]. The generalization of the noisy-OR by Srinivas is different to this proposal, and is intended for a different application. This generalization is for circuits (or other such devices) which can be either functional or non-functional. In the case of medicine, findings can take on values such as {absent, mild, moderate, severe}, in which case the binary generalization of Srinivas is insufficient to deal with arbitrary n-ary variables. The noisy-MAX generalization in [Diez 1993] is virtually identical to the one described here, and we have derived our noisy-MAX independent of that in [Diez 1993]. Also, the proposal in [Diez 1993] is described within the context of learning models for OR-gates, and its application to inference in Bayesian networks is not directly apparent.

To our knowledge, there is no other tool which allows dynamic selection of subsets of Bayesian networks. There are several graphical tools for creating Bayesian networks, including IDEAL Edit, Ergo, Hugin [Andersen, Olesen et al. 1989], and Demos [Morgan, Henrion et al. 1987]. But these tools do not provide dynamic network layout and do not have features aimed at knowledge engineering large BNs.

## 7  CONCLUSION

In this paper we have presented several methods for representing, and a software tool for managing, large BNs based on our experience with the CPCS BN. The noisy-MAX is a generalization of the noisy-OR gate for multivalued variables which reduces the complexity of the knowledge acquisition task and storage requirements for a network. Leak probabilities are used in the CPCS BN to model causes other than those explicitly modeled in the network.

Based on the causal independence assumptions of the noisy-MAX, and the use of leak probabilities we have developed Netview, a tool for visualizing BNs based on the dynamic layout of subnetworks, and which also provides basic version control for editing networks. The creation of subnetworks allows for more efficient knowledge engineering, and for easier verification of the BN. We describe a technique for updating leak probabilities based on the excluded parents of a node in subnetworks.

Recent advances in creating BNs from pre-existing data or knowledge bases will result in networks that are larger and more complex than those created manually. We believe that the techniques described in this paper are important to facilitate the management and verification of such networks.

## Acknowledgments

This work was supported by NSF Grant Project IRI-9120330, and by computing resources provided by the Stanford University CAMIS project, which is funded under grant number LM05305 from the National Library of Medicine of the National Institutes of Health.

The authors would like to thank K. C. Chang and R. Fung for their graph layout algorithm on which NetView's layout method is based, and R.Miller for access to the CPCS knowledge base.

## References

Andersen, S. K., Olesen, K. G., et al. (1989). HUGIN-a shell for building Bayesian belief universes for expert systems. *IJCAI-89 Proceedings of the Eleventh International Joint Conference on Artificial Intelligence*, Detroit, MI, USA, 20-25 Aug., pages 1080-5. Morgan Kaufmann, Palo Alto, CA, USA.

Cooper, G. F. (1986). A diagnostic method that uses causal knowledge and linear programming in the application of Bayes' formula. *Computer Methods and Programs in Biomedicine*, 22:223-237.

Diez, F. J. (1993). Parameter adjustment in bayes networks: The generalization of the noisy-OR gate. *Uncertainty in Artificial Intelligence,* Washington D.C., pages 99-105. Morgan Kaufmann.

Henrion, M. (1988). Practical issues in constructing a Bayes' belief network. *Uncertainty in Artificial Intelligence 3*. Amsterdam, North Holland. 132-139.

Middleton, B., Shwe, M. A., et al. (1991). Probabilistic diagnosis using a reformulation of the Internist-1/QMR knowledge base-II. Evaluation of diagnostic performance. *Methods of Information in Medicine*, 30:256-67.

Miller, R. A., Pople, H. E. J., et al. (1982). Internist-1: An experimental computer-based diagnostic consultant for general internal medicine. *N Engl J Med*, 307:468-476.

Minsky, M. (1975). A Framework for representing knowledge. *Psychology of Computer Vision*. Cambridge, MA, MIT Press.

Morgan, M. G., Henrion, M., et al. (1987). Demos: a computer aid for engineering-economic policy modeling and uncertainty analysis. *Large Scale Systems in Information and Decision Technologies*, 13(1):61-82.

Parker, R. C. and Miller, R. A. (1987). Using causal knowledge to create simulated patient cases: the CPCS project as an extension of Internist-1. *Proceedings of the Eleventh Annual Symposium on Computer Applications in Medical Care.,* Los Alamitos, CA, pages 473-480. IEEE Comp Soc Press.

Pearl, J. (1988). Probabilistic reasoning in intelligent systems. Morgan Kaufman, San Mateo, Ca.

Peng, Y. and Reggia, J. A. (1987). A probabilistic causal model for diagnostic problem solving - Part I: Integrating symbolic causal inference with numeric probabilistic inference. *IEEE Trans SMC*, SMC-17(2):146-162.

Provan, G. (1994). Causal Independence and Subnetwork Extraction in Large Belief Networks. Stanford University, CA, KSL-94-48, May.

Provan, G. M. (1993). Tradeoffs in constructing and evaluating temporal influence diagrams. *Uncertainty in Artificial Intelligence,* Washington D.C., pages 40-47. Morgan Kaufmann.

Reiter, R. (1978). On closed world data bases. *Logic and Data Bases*. New York, NY, Plenum. 300-310.

Shwe, M. A., Middleton, B., et al. (1991). Probabilistic diagnosis using a reformulation of the Internist-1/QMR knowledge base-I. The probabilistic model and inference algorithms. *Methods of Information in Medicine*, 30:241-55.

Srinivas, F. J. (1993). A Generalization of the noisy-OR model. *Uncertainty in Artificial Intelligence,* Washington D.C., pages 208-215. Morgan Kaufmann.

Srinivas, S. and Breese, J. S. (1989). IDEAL: Influence diagram evaluation and analysis in Lisp. Rockwell International Science Center, Palo Alto, Ca., May.